\documentclass[11pt]{article}
\usepackage{amsmath, amsthm, amssymb, mathrsfs, enumerate, graphicx, array, bm, polynom, xlop, color, hyperref, multirow}
\usepackage{soul}
\usepackage{enumitem}
\setlist[itemize]{leftmargin=*}
\usepackage[a4paper, total={15cm, 22.5cm}]{geometry}

\begin{document}
\title{Optimal minimal-perturbation university\\ timetabling with faculty preferences}
\author{
Jakob Kotas\thanks{Department of Mathematics, University of Portland, 5000 N Willamette Blvd, Portland OR 97203 USA; kotas@up.edu [Corresponding author]} \and Peter Pham\thanks{Shiley School of Engineering, University of Portland, 5000 N Willamette Blvd, Portland OR 97203 USA; phamp23@up.edu, koellman23@up.edu} \and Sam Koellmann\footnotemark[2]}
\maketitle
\begin{abstract}
In the university timetabling problem, sometimes additions or cancellations of course sections occur shortly before the beginning of the academic term, necessitating last-minute teaching staffing changes. We present a decision-making framework that both minimizes the number of course swaps, which are inconvenient to faculty members, and maximizes faculty members' preferences for times they wish to teach. The model is formulated as an integer linear program (ILP). Numerical simulations for a hypothetical mid-sized academic department are presented.
\end{abstract}

Keywords: scheduling, university timetabling, integer program, minimal perturbation

\section{Introduction}\label{sec:intro}

Scheduling and assignment problems have long been a focus of study in the operations research community. In particular, the problem of scheduling university courses has been investigated by a number of authors dating back at least to the 1970s.\cite{breslaw, shih} A number of models for university course scheduling, also known as timetabling, have been proposed. The overall problem contains subproblems including (a) determining time slots that courses may be offered during, (b) deciding which course is assigned to a particular room at a given time, (c) deciding which instructor is assigned to a given course, and so forth.\cite{carter} In this work we focus on the dynamic rescheduling problem that sometimes occurs after a timetable has been chosen, due to issues such as unexpected enrollment fluctuations. In this case, we wish to develop a new schedule which minimizes in some sense the number of course swaps, which are inconvenient to faculty members as well as students after registration has occurred; while at the same time taking into consideration faculty preferences for times they wish to teach.

In the scheduling literature, the reassignment problem is sometimes referred to as a ``minimal perturbation" problem (MPP). Bart\'{a}k, M\"{u}ller, Rudov\'{a}, and Murray were the first to study the MPP in the context of university course scheduling and have published several papers on the topic.\cite{bartak, muller, rudova} Their approach is called ``iterative forward search" which operates over feasible, but incomplete solutions (meaning some variables are unassigned). Their work builds off of El Sakkout, Richards, and Wallace\cite{elsakkout1,elsakkout2} who introduced the MPP for general dynamic programs. More recent work in high school timetabling has been that of Kingston\cite{kingston}, who describes an algorithm called ``polymorphic ejection chains," which repair an infeasibility while possibly creating a new one, in a successive fashion until the solution is feasible. Finally, Phillips, Walker, Ehrgott and Ryan\cite{phillips} minimize course swaps in the university timetabling problem by only searching for solutions in a neighborhood of the original (now-infeasible) solution, and expanding the neighborhood until a solution is found.

In contrast to the aforementioned works, the model we present here represents a new multiobjective approach, where swaps are minimized through explicit inclusion in the objective function. The benefits of this approach are several. First, ours is an exact solution method that does not rely on heuristics, so an optimal solution is guaranteed. Second, since both the minimizing of perturbations as well as a weighted sum of faculty time-preferences are considered in the objective function, we favor swaps that improve the times at which faculty teach. This also allows the decision-maker to choose the relative importance of these two objectives. Our formulation is an integer linear program (ILP) and thus can be solved with virtually any commonly-used optimization software package; there is no need for a problem-specific solution algorithm.

In this paper we borrow notation from a model put forth by Kumar.\cite{kumar} That model is also an ILP that schedules courses to time slots and assigns instructors to them.  The objective function is a linear weighted sum of faculty time-preferences. Note that Kumar's model is for the original timetabling assignment problem and \textit{not} the reassignment problem.


\section{Model}\label{sec:model}

\noindent\textbf{Decision variables}\\
\noindent\begin{tabular}{rcr p{9.75cm}}
    $P_{ijt}$ & = & 1 & if course $i$ is added to faculty $j$ in time slot $t$\\
     & = & -1 & if course $i$ is removed from faculty $j$ in time slot $t$\\
     & = & 0 & otherwise\\
\end{tabular}\\

Also $T_{ij}$ are decision variables which will be necessary to linearize the objective function; see below.\\

\noindent\textbf{Parameters}
\begin{itemize}
\item Obsolete schedule\\
\noindent\begin{tabular}{rcl}
    $X_{ijt}$ & = & 1 if course $i$ had been assigned to faculty $j$ in time slot $t$\\
    & = & 0 otherwise\\
\end{tabular}

The new schedule is $X_{ijt} + P_{ijt}$.

\item Faculty time-preference matrix\\
\noindent\begin{tabular}{rcc p{9.75cm}}
    $W_{jt}$ & $\geq$ & 0 & indicate faculty $j$'s preference for teaching in time slot $t$.
\end{tabular}

Higher scores correspond to more desirable times. 0 indicates unavailability during time slot.

\item Faculty time availability matrix\\
\noindent\begin{tabular}{rcc p{9.75cm}}
    $F_{jt}$ & = & 1 if $W_{jt} > 0$ & indicates faculty $j$ is available in time slot $t$\\
    & = & 0 if $W_{jt} = 0$ & otherwise\\
\end{tabular}

\item Course swap penalty matrix\\
\noindent\begin{tabular}{rcc p{9.75cm}}
    $\alpha_{ij}$ & $\geq$ & 0 & are penalty factors for making changes to course $i$ in faculty $j$'s schedule.\\
\end{tabular}

This penalty applies for either adding a course that was previously not in faculty $j$'s schedule, or for removing a course that was previously in faculty $j$'s schedule. Simply changing to a different section of an existing course incurs no penalty due to $\alpha_{ij}$. However note that a reward/penalty for changing the time slot to a more or less desirable time for faculty $i$ is handled via $W_{jt}$.

\item Course-time slot matrix\\
\noindent\begin{tabular}{rcl}
    $M_{it}$ & = & number of sections of course $i$ scheduled at time slot $t$\\
\end{tabular}

Note that due to this definition of $M_{it}$ differs from the Kumar model.

\item Faculty-course matrix\\
\noindent\begin{tabular}{rcl}
    $C_{ij}$ & = & 1 if course $i$ can be taught by faculty $j$\\
    & = & 0 otherwise\\
\end{tabular}

\item Teaching load\\
\noindent\begin{tabular}{lc p{10.25cm}}
    $N^+_{j}$ & = & maximum number of courses/credit hours that can be assigned to faculty $j$\\
    $N^-_{j}$ & = & minimum number of courses/credit hours that can be assigned to faculty $j$\\
    $H_i$ & = & number of courses/credit hours that course $i$ counts as
\end{tabular}

Our model accommodates faculty contracts on either a per-course or per-credit hour basis. If faculty member $j$ must teach an exact number, then $N^+_{j}=N^-_{j}$. Part-time faculty with a range on how many courses/credit hours (including zero) can also be accommodated.

\end{itemize}

\noindent\textbf{Objective function}

The objective function is the weighted sum of faculty time-preferences, plus the weighted sum of the absolute value of courses added and subtracted:
\[
\max \underbrace{\sum_j \sum_t \left( W_{jt} \sum_i P_{ijt} \right)}_{\textrm{faculty time preferences}} - \underbrace{\sum_i \sum_j \left( \alpha_{ij} \left|\sum_{t} P_{ijt}\right| \right)}_{\textrm{course swaps}}
\]

Where $|\cdot|$ is in the componentwise sense. Note that this objective function is not linear but can be made linear through the introduction of the new variables $T_{ij}$. We arrive at our final form:
\[
\boxed{\max \sum_j \sum_t \left( W_{jt} \sum_i P_{ijt} \right) - \sum_i \sum_j \alpha_{ij} T_{ij} }
\]
along with additional constraints labeled ``new variable constraints" below.\\

\noindent\textbf{Constraints}
\begin{itemize}
\item New variable constraints:

\hspace{1cm} $T_{ij} \geq \sum_t P_{ijt} \hspace{3mm}\forall i \hspace{3mm}\forall j$

\hspace{1cm} $T_{ij} \geq -\sum_t P_{ijt} \hspace{3mm}\forall i \hspace{3mm}\forall j$

\noindent These constraints are necessary to convert the $|\sum_t P_{ijt}|$ term in the objective function into linear form.

\item New schedule is binary:

\hspace{1cm} $X_{ijt}+P_{ijt} \geq 0 \hspace{3mm}\forall i \hspace{3mm}\forall j \hspace{3mm}\forall t$

\hspace{1cm} $X_{ijt}+P_{ijt} \leq 1 \hspace{3mm}\forall i \hspace{3mm}\forall j \hspace{3mm}\forall t$

\noindent This set of constraints ensures that you cannot add a course and time that a faculty member already has, nor can you remove a course and time that a faculty member did not have.

\item Assign all courses:

\hspace{1cm} $\sum_j (X_{ijt} + P_{ijt}) = M_{it}  \hspace{3mm}\forall t \hspace{3mm}\forall i$\\
\noindent These constraints ensure that every section of every course is assigned to a faculty member.

\item Faculty teach only courses from their choice list:

\hspace{1cm} $\sum_t (X_{ijt}+P_{ijt}) \leq C_{ij} \sum_t M_{it} \hspace{3mm}\forall i \hspace{3mm}\forall j$\\
\noindent These constraints ensure that faculty only teach courses they are able to teach.

\item Faculty teach only during their available times:

\hspace{1cm} $\sum_i (X_{ijt}+P_{ijt}) \leq F_{jt} \hspace{3mm}\forall j \hspace{3mm}\forall t$\\
\noindent These constraints ensure that faculty are not scheduled during times they are not available, and also that a faculty member may teach no more than one course simultaneously.

\item Teaching load requirements:

\hspace{1cm} $\sum_i H_i \sum_t (X_{ijt}+P_{ijt}) \leq N_j^+ \hspace{3mm}\forall j$

\hspace{1cm} $\sum_i H_i \sum_t (X_{ijt}+P_{ijt}) \geq N_j^- \hspace{3mm}\forall j$

\noindent These constraints ensure that each faculty member teaches the correct number of courses.

\item Avoid time slot conflicts:

\hspace{1cm} $\sum_i (X_{ijt'}+P_{ijt'}) + \sum_i (X_{ijt''}+P_{ijt''}) \leq 1 \hspace{3mm}\forall j \hspace{3mm}\forall t', t''$ that conflict\\
\noindent These constraints prevent a faculty from teaching at both time slots $t'$ and $t''$. This could be because $t'$ and $t''$ overlap or because we wish to prevent faculty from teaching \textit{both} early mornings and late evenings, or \textit{both} Mon/Wed/Fri and Tue/Thur, etc.

\item Decision variables are binary:

\hspace{1cm} $-1\leq P_{ijt}\leq 1$, $P_{ijt}$ integer

\hspace{1cm} $0\leq T_{ij}\leq 1$, $T_{ij}$ integer\\
This enforces that each element of the decision variable $P_{ijt} \in \{-1,0,1\}$ and $T_{ij} \in \{0,1\}$.

\end{itemize}

\section{Simulations}\label{sec:results}

In this section we present numerical simulations for a hypothetical department offering 57 sections of 17 different courses, with 13 full-time and 9 part-time faculty members, during 24 possible time slots. These values were based on the Fall 2020 semester schedule for the Mathematics Department at the University of Portland. All full-time faculty must teach 3 courses while part-time faculty may teach between 0 and 2 courses. Some pairs of time slots conflict with each other (time slots 8:10-9:05 Mon/Wed/Fri and 8:10-9:05 Mon/Tue/Wed/Fri are one such example.) We assume for simplicity that the $W_{jt}$ and $\alpha_{ij}$ matrices are all 1s, indicating that all faculty are equally satisfied teaching at any time and are equally inconvenienced by a course add or drop. $M_{it}$ represent actual sections scheduled for Fall 2020 at the University of Portland. For $C_{ij}$, an attempt was made to group together faculty based on their subdiscipline; for example, part-time faculty can teach any lower-division course, while pure mathematicians can teach lower-division courses plus pure upper-division courses, whereas applied mathematicians can teach lower-division courses plus applied upper-division courses. $X_{ijt}$ are a simulated feasible solution based on these constraints.

The IP involved 9350 variables and 39238 constraints. The problem was solved in Matlab R2016b, using the ``intlinprog" function in the Optimization Toolbox. Elapsed time to run the entire code, including reading in data from a spreadsheet, setting up constraints to pass to the solver, and displaying the output, was approximately 4 seconds on a single 2.8 GHz Intel Core i5 processor for each simulation that follows; the time for just the solution of the IP was approximately 2 seconds.\\

%

\noindent\textbf{Simulation 1: Removal from part-time faculty}

We remove one section of one course that is taught by a part-time faculty member in $X_{ijt}$. As expected, $P_{ijt} = -1$ for that course-faculty-time slot combination, and $P_{ijt} = 0$ for all other combinations indicating that no other changes occur.\\

\noindent\textbf{Simulation 2: Removal from full-time faculty}

We remove one section of one course that is taught by a full-time faculty member in $X_{ijt}$. Because full-time faculty must teach 3 courses, they are reassigned a course that had been taught by a part-time faculty member. Thus we have $P_{ijt} = -1$ in two places (the removed section and the section lost by the part-time faculty member) and $P_{ijt} = 1$ in one place (the section moved from part-time to full-time), and $P_{ijt} = 0$ elsewhere. We further note that the full-time faculty member is swapped to another section of the same course, which incurs no penalty due to $\alpha_{ij}$.\\

\noindent\textbf{Simulation 3: Forcing a course swap}

In both previous simulations, our model was able to adjust so that no instructor was forced to take on a new course, as opposed to switching sections of an existing course. However, if we push the system far enough, this becomes inevitable. For example, 8 sections of MTH 201 are offered. We choose to cancel 3 sections assigned to part-time faculty and 1 section assigned to a full-time faculty. The full-time faculty is reassigned to MTH 112 as there are no available sections of MTH 201 left. This is the first simulation in which we have seen an instance of the course swap penalty being activated (i.e., $T_{ij}$ > 0 for some $i$, $j$.)\\

A summary of simulations 1-3 can be found in table \ref{tbl:sim_removals}.\\

\begin{table}[ht]
\begin{center}

Simulation 1.\\
\begin{tabular}{|c|c|c|c|c|c|}
\hline
\multicolumn{3}{|c|}{Sections removed}&\multicolumn{3}{|c|}{Sections added}\\
\hline
Course & Faculty & Time slot & Course & Faculty & Time slot\\
\hline
\multirow{2}{*}{MTH 161} & \multirow{2}{*}{G.T.} & MWF & \multicolumn{3}{|c}{\multirow{2}{*}{(None)}}\\
&&12:30-13:25&\multicolumn{3}{|c}{}\\
\cline{1-3}
\end{tabular}\\

Simulation 2.\\
\begin{tabular}{|c|c|c|c|c|c|}
\hline
\multicolumn{3}{|c|}{Sections removed}&\multicolumn{3}{|c|}{Sections added}\\
\hline
Course & Faculty & Time slot & Course & Faculty & Time slot\\
\hline
\multirow{2}{*}{MTH 161} & \multirow{2}{*}{P.N.} & MWF & \multirow{2}{*}{MTH 161} & \multirow{2}{*}{M.M.} & MWF \\
&&8:10-9:05&&&8:10-9:05\\
\hline
\multirow{2}{*}{MTH 161} & \multirow{2}{*}{M.M.} & MWF & \multicolumn{3}{|c}{}\\
&&14:40-15:35&\multicolumn{3}{|c}{}\\
\cline{1-3}
\end{tabular}

Simulation 3.\\
\begin{tabular}{|c|c|c|c|c|c|}
\hline
\multicolumn{3}{|c|}{Sections removed}&\multicolumn{3}{|c|}{Sections added}\\
\hline
Course & Faculty & Time slot & Course & Faculty & Time slot\\
\hline
\multirow{2}{*}{MTH 112} & \multirow{2}{*}{G.T.} & MWF & \multirow{2}{*}{MTH 112} & \multirow{2}{*}{T.B.} & MWF \\
&&8:10-9:05&&&8:10-9:05\\
\hline
\multirow{2}{*}{MTH 201} & \multirow{2}{*}{T.B.} & MWRF & \multicolumn{3}{|c}{}\\
&&9:15-10:10&\multicolumn{3}{|c}{}\\
\cline{1-3}
\multirow{2}{*}{MTH 201} & \multirow{2}{*}{M.G.} & MTWR & \multicolumn{3}{|c}{}\\
&&9:15-10:10&\multicolumn{3}{|c}{}\\
\cline{1-3}
\multirow{2}{*}{MTH 201} & \multirow{2}{*}{D.K.} & MTWR & \multicolumn{3}{|c}{}\\
&&8:10-9:05&\multicolumn{3}{|c}{}\\
\cline{1-3}
\multirow{2}{*}{MTH 201} & \multirow{2}{*}{R.L.} & MTWR & \multicolumn{3}{|c}{}\\
&&12:30-13:25&\multicolumn{3}{|c}{}\\
\cline{1-3}
\end{tabular}

\end{center}
\caption{Simulations of courses removed. In simulation 1, a section is removed from part-time faculty member G.T. In simulation 2, a section is removed from full-time faculty member M.M, so a swap occurs of a different section from part-time faculty member P.N. to M.M. In simulation 3, 4 sections of MTH 201 are removed, one of which had been assigned to full-time faculty member T.B., so T.B. is reassigned to part-time faculty member G.T.'s MTH 112 section.}\label{tbl:sim_removals}
\end{table}

\section{Conclusion}\label{sec:conclusion}
We have developed a decision-making framework for last-minute teaching staffing changes when courses are added or removed for an academic department. This framework minimizes swaps in such a way that it penalizes changing a faculty member to a different course but not a different section of the same course. It also maximizes faculty members' preferences for times they wish to teach. The relative importance of these factors to each other as well as across various faculty members are handled through the weight matrices $W_{jt}$ and $\alpha_{ij}$. Our framework is easily and quickly solvable using off-the-shelf IP solvers. From the first author's experience teaching in an academic department, last-minute timetable changes are stressful to all involved; our hope is that this framework provides a tool that decision-makers can use to quickly, optimally, and objectively reshuffle teaching assignments when necessary.

\section{Acknowledgments}\label{sec:acknowledge}

This work was supported by a Summer Undergraduate Research Experience grant from the University of Portland College of Arts \& Sciences.

\bibliographystyle{plain}
\bibliography{lmrp.bib}

\end{document}